\documentclass[10pt,twocolumn,letterpaper]{article}
\usepackage{cvpr}
\usepackage{times}
\usepackage{epsfig}
\usepackage{graphicx}
\usepackage{amsmath}
\usepackage{amssymb}
\usepackage{bm}
\usepackage{enumerate}
\usepackage{color}
\usepackage{multirow}
\usepackage{array}
\usepackage{bigstrut}
\usepackage{algorithmic}
\usepackage{algorithm}
\usepackage{arydshln}
\usepackage{subfig}

\graphicspath{{figures/}}

\usepackage[pagebackref=true,breaklinks=true,letterpaper=true,colorlinks,bookmarks=false]{hyperref}

\newcommand{\comment}[1]{}
\newcommand{\deflen}[2]{%
    \expandafter\newlength\csname #1\endcsname
    \expandafter\setlength\csname #1\endcsname{#2}%
}

\def\cvprPaperID{2420} 

\ifcvprfinal\fi
\begin{document}

\title{Confidence Inference for Focused Learning in Stereo Matching}

\author{Ruichao Xiao \and Wenxiu Sun \and Chengxi Yang \and
SenseTime Research\\
{\tt\small \{xiaoruichao, sunwenxiu, yangchengxi\}@sensetime.com}
}
\maketitle

\begin{abstract}
In this paper, we present confidence inference approach in an unsupervised way in stereo matching. Deep Neural Networks (DNNs) have recently been achieving state-of-the-art performance. However, it is often hard to tell whether the trained model was making sensible predictions or just guessing at random. 
To address this problem, we start from a probabilistic interpretation of the $L_1$ loss used in stereo matching, which inherently assumes an independent and identical ({\it aka  i.i.d.}) Laplacian distribution. We show that with the newly introduced dense confidence map, the identical assumption is relaxed. Intuitively, the variance in the Laplacian distribution is large for low confident pixels while small for high-confidence pixels. In practice, the network learns to \emph{attenuate} low-confidence pixels ({e.g.,} noisy input, occlusions, featureless regions) and \emph{focus} on high-confidence pixels. Moreover, it can be observed from experiments that the focused learning is very helpful in finding a better convergence state of the trained model, reducing over-fitting on a given dataset.
\end{abstract}


\section{Introduction}
\label{sec:intro}
Understanding the confidence level of the prediction is a critical part in deep learning. Since there are numerous parameters in a model, it is often hard to tell whether the trained model was making sensible predictions or just guessing at random. With recent engineering advances in the field of machine learning, systems that were only applied to toy data are now being deployed in real-life settings. Among these settings are scenarios in which control is handed-over to automated systems, including automated decision making, recommendation systems in the medical domain, autonomous control of drones and self driving cars, as well as control of critical systems. 

In particular, for deep regression tasks, state-of-the-art performances~\cite{pang2017cascade,iresnet} are achieved with Deep Neural Networks (DNNs). However, without confidence measures, the predictions are often assumed to be accurate, which is not always the case. Taken the stereo matching problem as an example, the best trained model on KITTI Stereo 2015~\cite{Menze2015CVPR} can reach up to an error rate of $2.32\%$ measured by the percentage of outliers over all ground truth pixels. But still, there are $2.32\%$ wrongly predicted pixels. If those pixels appear at some critical objects, such as a thin rail, it could be dangerous for depth-assisted obstacle avoidance systems or advanced driver-assistance systems. Moreover, worse or unreasonable predictions can be observed if tested on a different dataset or on degraded inputs.   

While confidence can be manually designed from a list of hand-crafted rules for classical methods, it is not very straight-forward to design such rules in deep learning. Both noisy data and unsuited model can lead to the degraded performance of stereo matching with deep learning. Particularly, noisy data includes the falsely labeled ground-truth, the noise corrupted input, or the ill-posed regions (occluded regions, repeated patterns, featureless regions, {\it etc.}). Unsuited model includes not-well-designed network structure, not-well-trained model (under-fitting or over-fitting), {\it etc.}.


\begin{figure*}[t]
\centering
    \includegraphics[width=\linewidth]{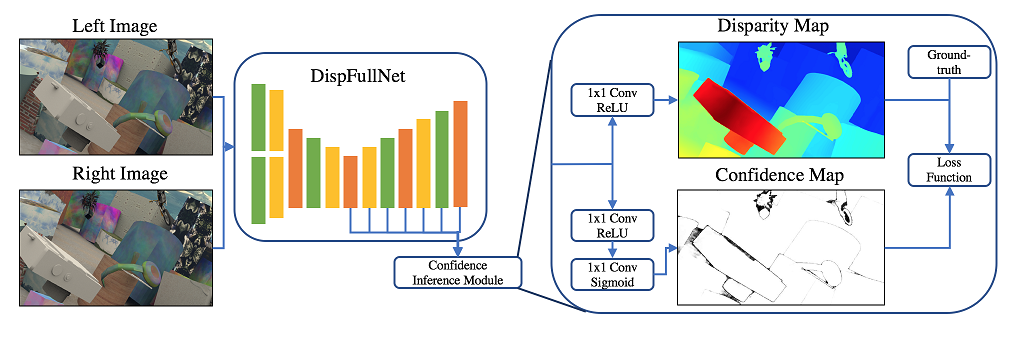}
\caption{The proposed network architecture for confidence learning with a new loss function. The network learns to \emph{attenuate} low confident pixels ({e.g.,} noisy input, occlusions, featureless regions), and \emph{focus} more on high confident pixels.}
\label{fig:structure}
\end{figure*}

In this paper, we start from a probabilistic interpretation of the $L_1$ loss used in stereo matching, which inherently assumes an independent and identical ({\it aka  i.i.d.}) Laplacian distribution. Intuitively, there is a strong correlation between the variance in the Laplacian distribution and the confidence level for an arbitrary pixel. That is, the variance of the Laplacian distribution is large for low confident pixels while small for high confident pixels. By introducing the confidence as an additional variable with certain distribution parameterized by $\gamma$ in our new formulation, we show that the identical distribution assumption is relaxed. Interestingly, this leads to a new loss function, where 1) the original $L_1$ loss is attenuated at low confident regions, reducing their influence to other pixels during back-propagation; 2) the confidence is penalized for being low with the confidence regularization term. The network structure is shown in Figure~\ref{fig:structure}.
In practice, the network learns to \emph{attenuate} low confident pixels ({e.g.,} noisy input, occlusions, featureless regions), as the attenuated $L_1$ loss produces relatively low cost. Meanwhile, the network \emph{focuses} more on high confident pixels, as the confidence regularization term trade-offs the cost. Moreover, by deploying the network to other stereo matching dataset, it can be observed from experiments that the focused learning is very helpful in finding a better convergence state of the trained model, reducing over-fitting on a given dataset. Different with~\cite{kendall2017uncertainties} which implicitly treats the confidence distribution as uniform distribution, a special case when $\gamma = 0$, we also study how different confidence distributions would affect the performance of a model. 
The main contributions of this paper can be summarized as follows,
\begin{itemize}
 \item We propose a confidence inference module which does not require ground-truth confidence labels. The inferred confidence has its physical meaning, which can be employed to facilitate the decision-making or the post-processing tasks.
  \item We show that with the newly introduced confidence, the identical Laplacian distribution assumption is relaxed. Particularly, the variance of the Laplacian distribution is large for low confident pixels while small for high confident pixels.
  \item We observe from experiments that the proposed method is very helpful in finding a better convergence state of the trained model, reducing over-fitting on a given dataset.
\end{itemize}

This paper is organized as follows. 
Related works are reviewed in Section\;\ref{sec:related}. 
We then present our confidence inference approach in Section\;\ref{sec:method}. 
Section\;\ref{sec:results} presents the experimentation and Section\;\ref{sec:conclude} concludes our work.

\section{Related Works}
\label{sec:related}
A taxonomy of confidence measures have been proposed by Hu and Mordohai~\cite{hu2012quantitative}. They categorize 17 confidence prediction methods in stereo matching into 6 categories according to the cues exploited in confidence prediction. Moreover, they proposed an effective metric to assess the effectiveness of confidence prediction based on Area Under the Curve (AUC) analysis. Most recently, Poggi {\it et  al.}~\cite{poggi2017quantitative} proposed an updated review and quantitative evaluation of 52 state-of-the-art confidence prediction methods, including some of the deep learning methods. In this section, we review the most related works from a different viewpoint based on their methodology.

\subsection{Confidence prediction with hand-crafted cues} In this category, the confidence is predicted by some defined metric based on ones' expert knowledge. Egnal {\it et al.}~\cite{egnal2004stereo} proposed to use the negative cost as a simple confidence measure, so that large values correspond to higher confidence. They also proposed to utilize the shape of the cost curve around the minimum as an indication of confidence. Winner margin is proposed by Scharstein and Szeliski~\cite{scharstein1998stereo} which compares two smallest local minima of the matching cost. Left-right consistency proposed by Egnal {\it et al.}~\cite{egnal2004stereo} is defined as the absolute difference between the disparity in the left image and the corresponding disparity in the right image. Seki and Pollefeys~\cite{seki2016patch} proposed to infer a confidence measure by processing features extracted from the left and right disparity
maps. Distinctiveness~\cite{manduchi1999distinctiveness}  can also be used as a cue for confidence prediction as did in Yoon {\it et al.}~\cite{yoon2008distinctive}. Their assumption is that distinctive points ({\it e.g.,} edges and corners) are less likely to be falsely matched between reference and target images. Xu {\it et al.}~\cite{xu2008stereo} treat occlusion regions as low confident regions and exclude them in the stereo matching process for better performance. 

\subsection{Confidence leanring with inference learning} 
The hand-crafted cues applies mostly in classical stereo matching methods for confidence prediction. However, it is not an easy task to apply such rules directly in deep learning. In contrast, some regression or classification methods for confidence prediction are being proposed. In the perspective of machine learning, Haeusler {\it et al.}~\cite{haeusler2013ensemble} proposed a confidence prediction method by classifying feature vectors made by 23 hand-crafted confidence measures. Similarly,  Park~{\it et al.}~\cite{park2015leveraging} trained regression forests to predict the correctness (confidence) of a match by using selected confidence measures. Spyropoulos~{\it et al.}~\cite{spyropoulos2016correctness} proposed to predict confidence by using a random-forest to classify a set of hand-crafted features for each pixels. 
Meanwhile, in the domain of deep learning, Poggi and Mattoccia~\cite{poggi2016learning} classified the predicted disparity map to confident patches and un-confident patches. Similarly, Shaked{\it et al.}~\cite{shaked2017improved} also proposed a reflective loss which is used to classify prediction of disparity to binary labels. Gurevich~{\it et al. }~\cite{gurevich2017learning} simultaneously trained two neural networks with a joint loss function. One of the networks performs predictions and the other simultaneously quantifies the uncertainty of predictions by estimating the locally averaged loss of the first one. Ummenhofer {\it et al.}~\cite{ummenhofer2017demon} proposed a supervised approach to regress a confidence, where the supervision is generated during training based on the disparity prediction error. 
\subsection{Confidence prediction with probabilistic modeling} Another category of methods address the problem from the probabilistic point of view. Based on the work of Zhang and Shan~\cite{zhang2000progressive} which calculates costs on a similarity function by treating the value assigned to each potential disparity as a probability for the disparity. Hu and Mordohai~\cite{hu2012quantitative} achieved a simple confidence measure by normalized cost values, as they do not attempt to convert cost to an exact confidence. Recently, Kendall and Gal~\cite{kendall2017uncertainties} proposed to learn epistemic uncertainty and aleatoric uncertainty for Bayesian Neural Networks, which relates to confidence measures. 

Our proposed method is a deep learning approach, and belongs to confidence prediction methods with probabilistic modeling. In particular, we assume that the predicted disparity have large variance when it is less confident, and smaller variance when it is more confident. By maximizing the likelihood of the predicted disparity and confidence, we derive a simple yet very effective loss to simultaneously learn disparity and its confidence.

\section{Methodology}
\label{sec:method}
Deep learning is not perfect. It is often observed that a well-trained model on one dataset may fail easily on another. In order to make ourselves or the control systems aware of when and where the deployed model would fail, we intend to infer for stereo matching a dense confidence map for the predicted disparity. 

\subsection{What is Confidence?} \label{sec:what_is_confidence}
Predicting a confidence is not straight-forward, given the fact that
a) the definition of confidence is subjective; b) there is no ground-truth available for a supervised confidence training. Thus, before estimating a dense confidence map, let's first make it clear what is the targeted confidence that we want to generate,
\begin{enumerate}
\item The confidence should be high for correct regions and low for error regions.
\item The confidence values are in a range of $[0, 1]$.
\end{enumerate}

\subsection{Probabilistic Interpretation} 
To infer the confidence, let us start from a probabilistic interpretation of the $L_1$ loss used in stereo matching~\cite{mayer2016large,pang2017cascade}, which inherently assumes an independent and identical ({\it aka  i.i.d.}) Laplacian distribution. 
Let $\mathbf{x}=\{x_1, x_2, ..., x_N\}$ be the input of the network and $\mathbf{y}=\{y_1, y_2, ..., y_N\}$ be the predicted disparity map, where $N$ is the number of pixels in the input images. Usually, each pixels has a corresponding disparity value.
Let $\mathbf{w}$ be the model parameter. The optimal model parameter is found by maximizing the following likelihood function,
\begin{equation} \label{eq:yw}
\prod_i^N P(y_i,\mathbf{w}|\mathbf{x}) = \prod_i^N P(y_i | \mathbf{w}, \mathbf{x}) P(\mathbf{w} | \mathbf{x} )
\end{equation}
Assuming the observed disparity values follow an identical Laplacian distribution,
\begin{equation}\label{eq:y}
 P(y_i | \mathbf{w}, \mathbf{x}) \propto \frac{1}{2b} e^{-\frac{|y_i - f_i^{\mathbf{w}}(\mathbf{x})|}{b}}
\end{equation}
and the model parameter is independent of the input $\mathbf{x}$ and follows another Laplacian distribution with zero-mean and variance equals to one,
\begin{equation}\label{eq:w}
 P(\mathbf{w} |  \mathbf{x}) \propto e^{-| \mathbf{w}|}
\end{equation}
Substituting (\ref{eq:y}) and (\ref{eq:w}) into (\ref{eq:yw}), and take the negative log-likelihood, 

\begin{eqnarray}
\lefteqn{\sum_i^N  -\log P(y_i | \mathbf{w}, \mathbf{x})} \\
& &= \sum_i^N -\log P(y_i | \mathbf{w}, \mathbf{x}) - \log P(\mathbf{w} | \mathbf{x})\\
& &\propto \sum_i^N  \frac{|y_i - f_i^{\mathbf{w}}(\mathbf{x})|}{b} + \log 2b  + \lambda |\mathbf{w}|
\end{eqnarray}
where $\lambda$ is a scalar, corresponding to the weight decay during the process of model training. Conventionally, we add $\frac{1}{N}$ as a normalizer to re-scale the loss function and gradient. Now the loss function is defined as,
\begin{equation} \label{eq:l1}
\mathcal{L} = \frac{1}{N}\sum_i^N \underbrace{\frac{|y_i - f_i^{\mathbf{w}}(\mathbf{x})|}{b}}_{L_1 \text{ loss}}  +  \underbrace{\lambda|\mathbf{w}|}_{\text{weight regularization}} + \underbrace{\log 2b}_{\text{const}}
\end{equation}
which corresponds to the commonly used $L_1$ loss function in stereo matching, with weight regularization.


\subsection{Confidence Learning} \label{sec:conf_learning}
Let us denote the confidence map as $\mathbf{c} = \{c_1, c_2, ..., c_N\}$ where $c_i \in [0, 1]$ as a new random variable. Based on the confidence properties discussed in Section~\ref{sec:what_is_confidence} that the confidence should be high for correct regions and low for error regions. We assume that the variance in the Laplacian distribution is large for low confident pixels and small for high confident pixels. For simplicity here, we set the variance $b_i$ as a linearly decreasing function of $c_i$,
\begin{equation} \label{eq:var}
b_i = f(c_i) = -k c_i + a
\end{equation}
where $k$ and $a$ are two positive constants satisfying $a 
\geq k + 1$, such that the variance always satisfies $b_i \geq 1$. With the newly introduced confidence $\mathbf{c}$, the likelihood function in~(\ref{eq:yw}) changes to,
\begin{eqnarray} \label{eq:ywc}
\lefteqn{\prod_i^N P(y_i, c_i, \mathbf{w} |\mathbf{x}) = } \\
& & \prod_i^N P(y_i |  c_i, \mathbf{w}, \mathbf{x}) P(c_i | \mathbf{w}, \mathbf{x}) P(\mathbf{w} | \mathbf{x})
\end{eqnarray}
Intuitively, it is favored that the confidence follows a non-decreasing distribution. We will elaborate it more in Sec.~\ref{discussion}. For simplicity and here, let us define the probability density function of confidence as the following, though other non-decreasing function also applies.
\begin{equation} \label{eq:c}
P(c_i | \mathbf{w}, \mathbf{x}) \propto c_i^{\gamma}
\end{equation}
where $\gamma \geq 0$.
Take the negative log-likelihood of ~(\ref{eq:ywc}), 
\begin{eqnarray}\label{eq:lognewloss}
\lefteqn{\sum_i^N -\log P(y_i, c_i, \mathbf{w} |\mathbf{x})} \\
&=& \sum_i^N -\log P(y_i | c_i, \mathbf{w}, \mathbf{x}) - \log P(c_i |\mathbf{w}, \mathbf{x})  \\
& &- \log P(\mathbf{w} | \mathbf{x})  \\
&\propto& \sum_i^N \{\frac{|y_i -f^\mathbf{w}(\mathbf{x})|}{-k c_i + a} + \log 2(-kc_i+a) -\\
& & \gamma \log c_i\} + \lambda |\mathbf{w}| 
\end{eqnarray}
Similarly, we multiply $\frac{1}{N}$ as normalizer and the new loss function is defined as,
\begin{eqnarray}\label{eq:newloss}
\hat{\mathcal{L}}(\mathbf{w}) &=& \frac{1}{N}\sum_i^N \{\underbrace{\frac{|y_i -
f^\mathbf{w}(\mathbf{x})|}{-k c_i + a}}_{\text{Focused }L_1\text{ loss}} + \\
& &\underbrace{ \log (-kc_i+a)  - \gamma \log c_i}_{\text{confidence regularization}}\} +\\
& & \underbrace{ \lambda |\mathbf{w}| }_{\text{weight regularization}} + \underbrace{C }_{\text{constant}}
\end{eqnarray}
Compared to~(\ref{eq:l1}), there are two key differences. First, the $L_1$ loss changes to \emph{focused} $L_1$ loss. For high confident pixels, where $f(c_i) = 1$, the loss is unchanged. For low confident pixels, where $f(c_i) > 1$, the loss is attenuated. Therefore, the first term focuses more on confident pixels. Second, a new regularization term called \emph{confidence regularization} is introduced, which penalizes low confidences. Another The loss function is fully deferential-able respects to $c_i$, thus the confidence will be learned inherently

In practice, as shown in Figure~\ref{fig:structure}, we add at the end of the network an additional convolution followed by a Sigmoid layer to bound the output between $0$ and $1$. The new loss function is used instead of the default $L_1$ loss. 

\subsection{Discussion on the formulation}\label{discussion}
Kendall~{\it et. al,}~\cite{kendall2017uncertainties} discussed this problem from a different point of view and resulted in similar loss formulations. However, there is a key difference that the formulation in~\cite{kendall2017uncertainties} inherently assumes that the confidence follows a uniform distribution, corresponding to a special case with $\gamma = 0$ in our formulation.

\begin{figure}[!t]
    \centering
\subfloat[$\gamma=0$]
{\includegraphics[width=0.8\linewidth]{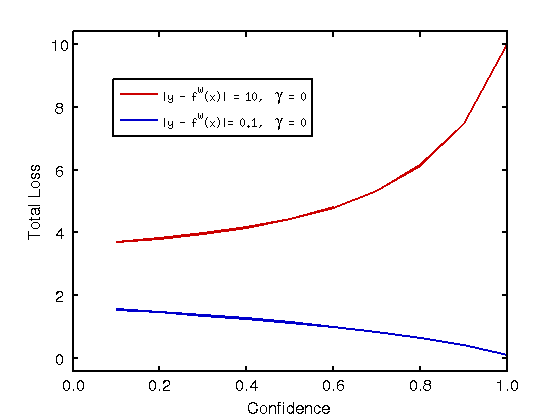}}\hspace{0pt} 
\subfloat[$\gamma=1$]
{\includegraphics[width=0.8\linewidth]{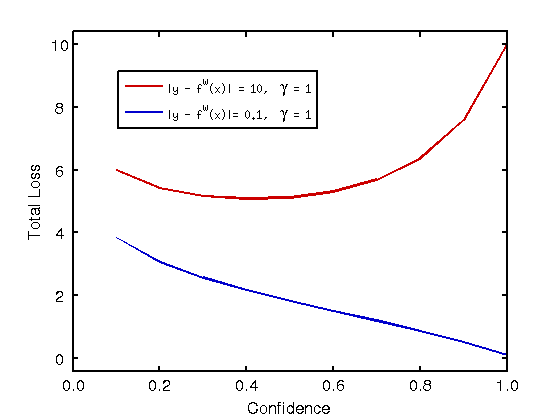}}\hspace{0pt}
 \caption{The optimal solution is achieved with high confidence for accurate disparity predictions (blue curve) and low confidence at regions of in-accurate disparity predictions (red curve). }
 \label{fig:loss_illustration}
\end{figure}

Note that with the introduction of confidence, the $L_1$ loss defined in~(\ref{eq:newloss}) is re-weighted by the confidence. 
The optimal solution is achieved with high confidence at regions of accurate predictions and low confidence at regions of in-accurate predictions. For easy understanding, we plot sample loss curves in Figure~\ref{fig:loss_illustration}, where the horizontal axis is the confidence, the vertical axis is the total loss, the two lines are two loss curves with $|y_i -f^\mathbf{w}(\mathbf{x})| = 10$ for red curve, and $|y_i - f^\mathbf{w}(\mathbf{x})| = 0.1$ for blue curve. Let us look at Figure~\ref{fig:loss_illustration}(a) first, where $\gamma=0$. Clearly, for the red curve with large prediction error, the optimal loss is achieved at confidence value $c=0$. For the blue curve with small prediction error, the optimal loss is achieved at confidence value $c=1$. Next, let us look at Figure~\ref{fig:loss_illustration}(b), where $\gamma=1$. For the red curve with the same large error as that in Figure~\ref{fig:loss_illustration}(a), the optimal confidence is at $c=0.42$. In this case, the network did not fully give up on these hard regions. It is good for achieving a good performance on the training dataset, but may result in an over-fitted model if the network is impossible to infer correct values at those regions. Again, it is important to stress that the confidence learning is not an ad-hoc construction, but a consequence of maximum likelihood estimation (MLE).

\section{Experiments}
\label{sec:results}
We present experiment results of our confidence learning approach on the stereo matching task. Firstly, we conduct ablation-study to verify the effectiveness of the proposed confidence learning on Flyingthings3D dataset~\cite{mayer2016large}. Then, we show that the confidence learning approach can help obtain a model with better generalization ability, when deploying the pre-trained model on a synthetic dataset to a real-world dataset with different characteristics.
Finally, we test our method on KITTI Stereo 2015 dateaset~\cite{Menze2015CVPR} to validate its effectiveness on the real-world scenario.
\subsection{Evaluation Metrics}
To evaluate the confidence map, we adopt the Receiver Operating Characteristic (ROC) curve and its Area Under the Curve (AUC) proposed by Hu~{\it et al. } ~\cite{hu2012quantitative}.
Specifically, we calculate the disparity error rate of the top $i\%$ confidence pixels ($i = 5, 10, 15, ..., 100$),
where error rate is defined as the percentage of pixels with disparity error larger than $\theta$ pixels. Particularly, in our experiment, we set $\theta=1$ as suggested in ~\cite{hu2012quantitative} when plotting the ROC curve. The overall ROC curve is plotted with its values calculated from the entire dataset. AUC calculates the area under the ROC curve. Lower AUC indicates better ability of the confidence map to identify correct disparity pixels. Additionally, according to ~\cite{hu2012quantitative}, for a given error rate $\epsilon$ at full density, there exists an optimal AUC, which is calculated as,
\begin{equation}
AUC_{opt} = \epsilon + (1-\epsilon)\ln(1-\epsilon) 
\end{equation}
Since the error rate at full density is different among the compared methods, it is not fair enough to compare their AUC scores directly. As done in~\cite{hu2012quantitative}, we further calculate the ratio between the average AUC score and the optimal AUC score to compare the confidence generated by different methods. 

To evaluate the disparity map, the first metric that we use is the commonly adopted end-point-error (EPE),  which is calculated as the absolute value between the predicted disparity and the ground-truth disparity. The second metric to evaluate the disparity map is the error rate, which is calculated as the percentage of disparities with their EPE larger than $t$ pixels. Note that when $t=\theta$, the error rate is that in the ROC curve at full density.

\subsection{Ablation Studies}\label{sub:abl}
We conducted several ablation experiments on FlyingThings3D~\cite{mayer2016large} dataset to justify the proposed approach. FlyingThings3D is a synthetic dataset containing $35,454$ training image pairs and $4,370$ test image pairs. 
Dense ground-truth disparities are provided for both training and test images. In addition, there are large portion of hard regions ({\it e.g.,} occlusions, featureless regions) in the dataset, which makes it efficient to evaluate the confidence inference approaches.

\begin{table*}[htbp]
\centering
\begin{tabular}{|c|c|c|c|c|c|c|c|c|}
\hline
\multicolumn{1}{|c|}{Methods} & \multicolumn{1}{c|}{$\gamma$} & $AUC_{opt}$ & $AUC_{avr}$ & Ratio & EPE & 1px Error& 3px Error& 5px Error\\ \hline
\multirow{6}{*}{Our Approach} & 0 & 0.0106 & 0.0478 & \textbf{0.2218} &1.4436&0.1420&0.06488&0.04609\\ \cline{2-9} 
 & 0.5 & 0.0104 & 0.0520 & 0.2003 & 1.4387 & 0.1409 & 0.06423 & 0.04547 \\ \cline{2-9} 
 & 1 & 0.0094 & 0.0588 & 0.1599 & \textbf{1.3945} & \textbf{0.1337}& \textbf{0.06077}& \textbf{0.04284}\\ \cline{2-9} 
 & 2 & 0.0113 &0.0612 & 0.1846 & 1.4322 & 0.1466 & 0.06418 & 0.04495 \\ \cline{2-9} 
 & 5 & 0.0118 & 0.0782 & 0.1509 & 1.4274 & 0.1499 & 0.06431 & 0.04467 \\ \hline
\multicolumn{2}{|c|}{DispFullNet~\cite{pang2017cascade}} & - & - & - & 1.6689 & 0.1960 & 0.07836 & 0.05235 \\ \hline
\end{tabular}
\vspace{8pt}
\caption{Ablation study results. $AUC_{avr}$ is the average $AUC$ of all images in the test dataset. Ratio is defined as the ratio of $AUC_{opt}$ over its corresponding $AUC_{avr}$. Larger Ratio indicates better confidence results.}
\label{tab:abl}
\end{table*}

\subsubsection{Model Architecture}
We use DispFullNet~\cite{pang2017cascade} as our baseline model.
The network architecture is similar to DispNetC~\cite{mayer2016large} but generates disparity map at the original resolution.
To estimate confidence map, we add a confidence inference module for each output scale to suit the multi-scale disparity training scheme in DispFullNet. as shown in Figure~\ref{fig:structure}. The output confidence map with largest resolution is used for evaluation.

All of our models are fine-tuned based on the pretrained weights given in ~\cite{pang2017cascade}, under the deep learning framework called CAFFE~\cite{jia2014caffe}. All models are optimized using the Adam method~\cite{Diederik2014ADAM} with $\beta_1$=0.9, $\beta_2=0.999$, and a batch size of 8. Multi-step learning rate is adopted during training. Initially, the learning rate was set to $5 * 10^{-5}$, and then reduced by half at the  $50$k-th, $100$k-th and $150$k-th iterations. The training was stopped at the $170$k-th iterations. The input images are randomly cropped to $768 \times 384$ during training and resize to $960 \times 576$ during testing.

\subsubsection{Parameters in Confidence Learning}\label{subsub:subj}
As mentioned in Section~\ref{sec:conf_learning}, we assume the confidence follows a non-decreasing distribution, as described in~(\ref{eq:c}). Particularly, we evaluate the performances by setting $\gamma = 0, 0.5, 1, 2$, and $5$. Moreover, we set $k=4$ and $a=5$ in (\ref{eq:var}) in our experiment. We choose this formulation for several reasons. Firstly, linear function is a simple function so we do not introduce other complexity in the experiments. Secondly, by restricting $b_i$ into a finite range, the stability is guaranteed during optimization. 

From the disparity error metrics (EPE, 1px Error, 3px Error, 5px Error) in Table~\ref{tab:abl}, it can be observed that all the best disparity is achieved when $\gamma=1$. A possible explanation is that a relatively large $\gamma$ prevents the aggressive diminishing of gradient on large error regions during back propagation.
Thus in terms of disparity estimation, 
we observe decreased error in EPE, and consequently, lower error rate. However, larger $\gamma$ (\textit{i}.\textit{e}. more weight on the confidence regularization) comes with a cost. It makes the model tends to assign high-confidence to reduce the confidence regularization and consequently, larger gradients may be included during back propagation.

From the confidence error metrics (ratio of $AUC_{opt}$ to $AUC_{avr}$) in Table~\ref{tab:abl}, we observe that the best confidence inference result is achieved when $\gamma=0$. This can be explained by the previous observation that small $\gamma$ can aggressively reject large error regions.

Finally, compared to the baseline model, our methods have significantly lower EPE and lower error rate with large margin (around relatively $20\%$ for all metrics), which suggests that our approach is able to help the model to reach a better convergence state.


\begin{figure}[htbp]
    \centering
\subfloat[(a)]{\includegraphics[width=0.24\linewidth]{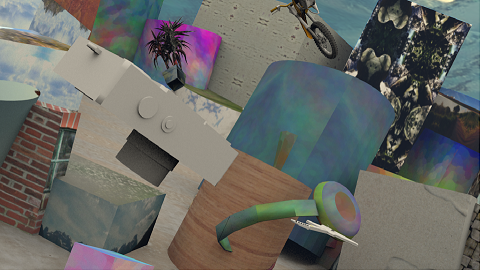}}\hspace{0.1pt} 
\subfloat[(b)]{\includegraphics[width=0.24\linewidth]{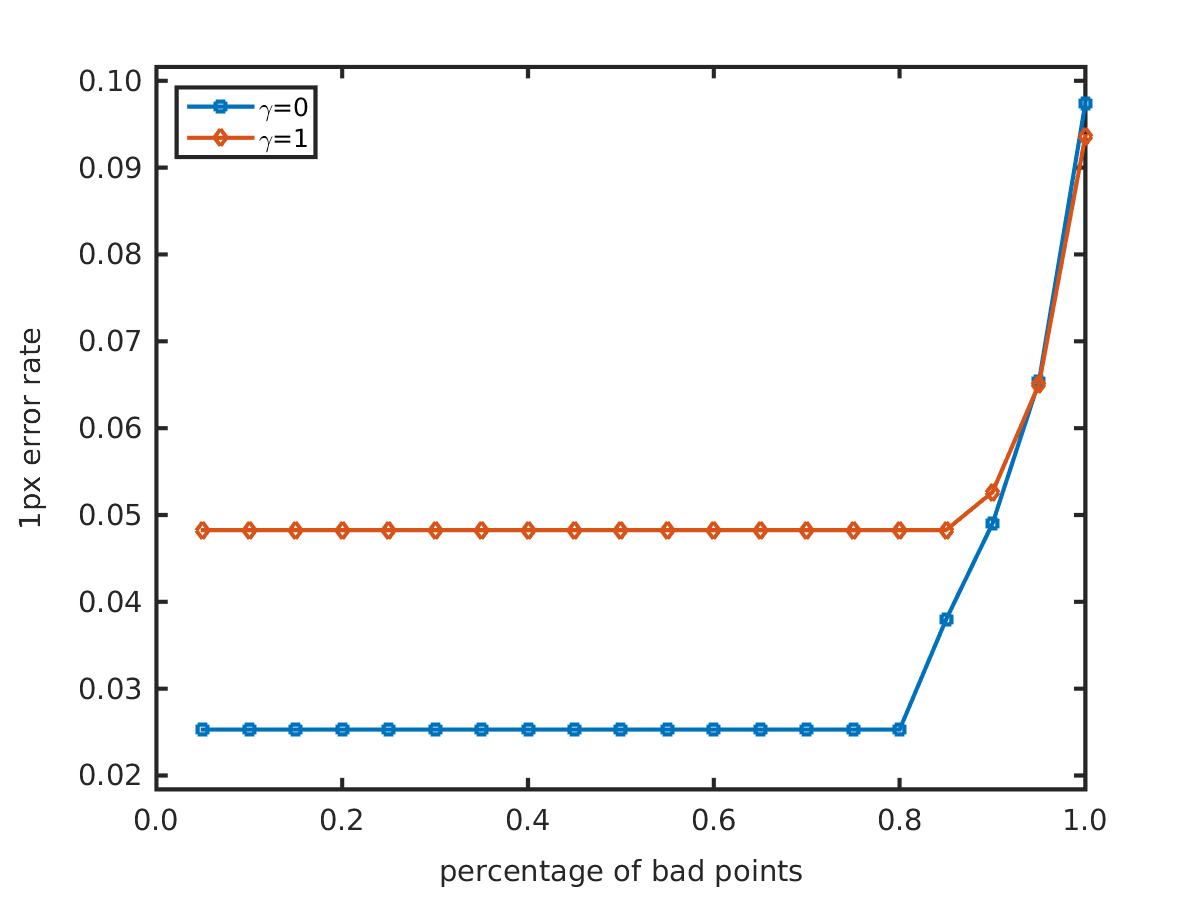}}\hspace{0.1pt}
\subfloat[(c)]{\includegraphics[width=0.24\linewidth]{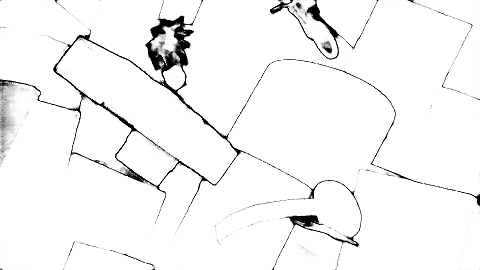}}\hspace{0.1pt}
\subfloat[(d)]{\includegraphics[width=0.24\linewidth]{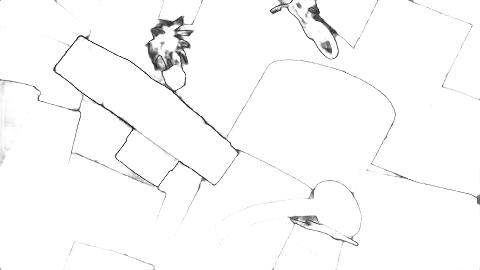}}\hspace{0.1pt}
\subfloat[(e)]{\includegraphics[width=0.24\linewidth]{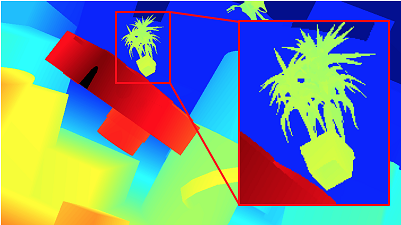}}\hspace{0.1pt} 
\subfloat[(f)]{\includegraphics[width=0.24\linewidth]{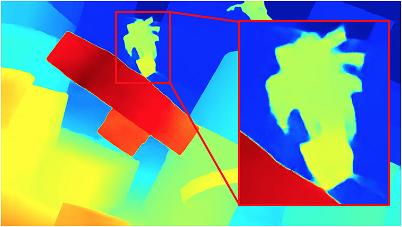}}\hspace{0.1pt}
\subfloat[(g)]{\includegraphics[width=0.24\linewidth]{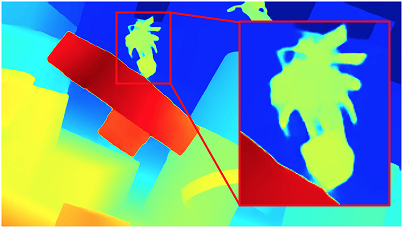}}\hspace{0.1pt}
\subfloat[(h)]{\includegraphics[width=0.24\linewidth]{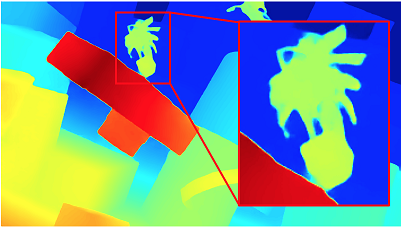}}\hspace{0.1pt} 
\vspace{-3pt}
 \caption{Some visual results on Flyingthings3D dataset. (a) Reference image; (b) ROC curve for $\gamma=0$ (blue) and $\gamma=1$ (red); (c)-(d): Confidence map for $\gamma=0$ and $\gamma=1$, respectively. Dark region means low confidence; (e)-(h): Disparity maps of ground-truth, DispFullNet, our method with $\gamma=0$, and our method with $\gamma=1$, respectively. Notice the fine details on the enlarged regions. Best view in color. 
}
\label{fig:fly}
\end{figure}

\begin{figure}[htbp]
    \centering
{\includegraphics[width=0.8\linewidth]{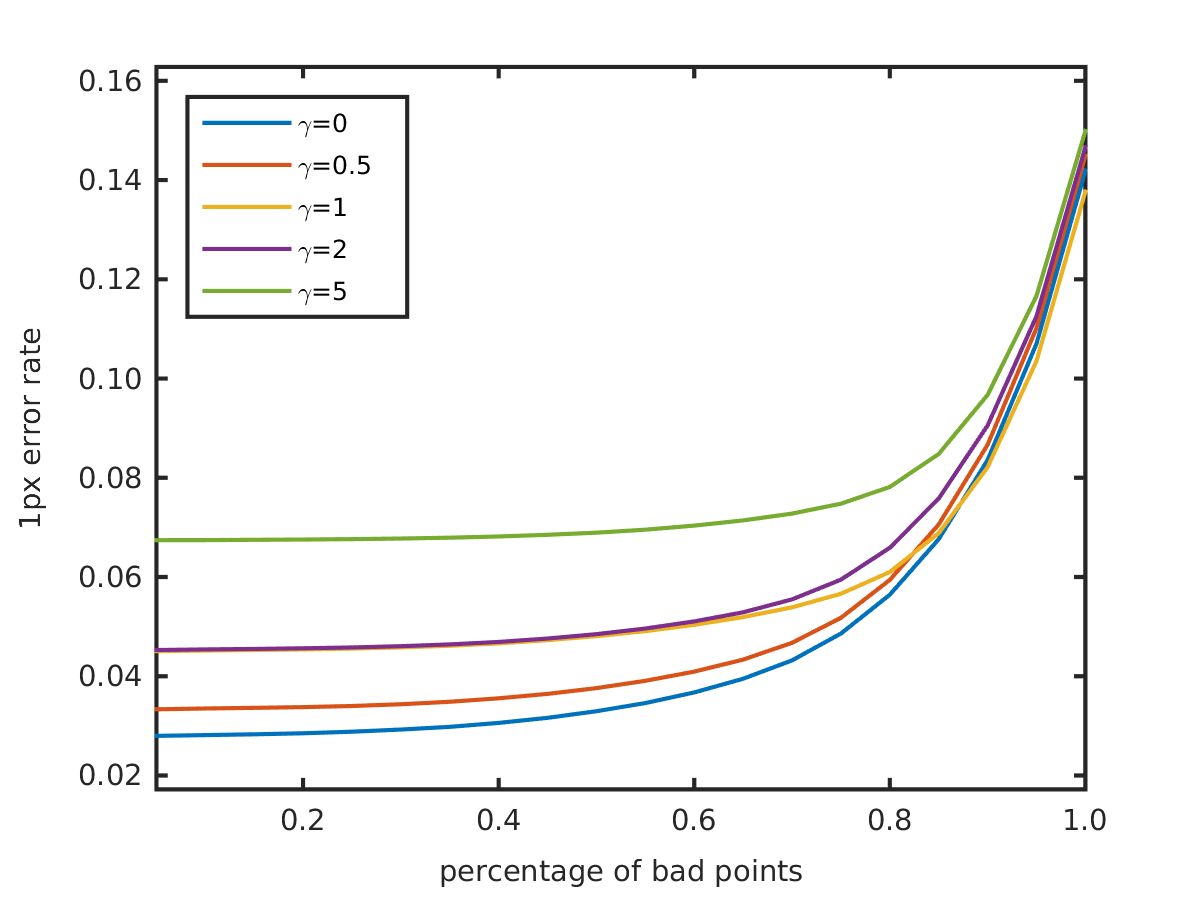}}\hspace{0.5pt} 
 \caption{ROC curve for $\gamma=0,0.5,1,2,5$ of the predicted confidence in the dataset of FlyingThings3D.
 }
 \label{fig:rocfly}
\end{figure}

\begin{table*}[htbp]
\centering
\begin{tabular}{|c|c|c|c|c|c|c|c|c|}
\hline
Method & $\gamma$ & $AUC_{opt}$ & $AUC_{avr}$ & ratio & EPE & 1px Error & 3px Error & 5px Error \\ \hline
\multirow{2}{*}{Ours} & 0 & 0.1103 & 0.2850 & \textbf{0.3870} & \textbf{6.6605} & \textbf{0.4366} & \textbf{0.2008} & \textbf{0.1449} \\ \cline{2-9} 
 & 1 & 0.1154 & 0.3067 & 0.3763 & 6.8425 & 0.4402 & 0.2047 & 0.1457 \\ \hline
\multicolumn{2}{|l|}{DispFullNet~\cite{pang2017cascade}} & - & - & - & 9.5944 & 0.4727 & 0.2686 & 0.2064 \\ \hline
\end{tabular}
\vspace{3pt}
\caption{Confidence and disparity evaluation results on Middleburry dataset.}
\label{tab:mid}
\end{table*}


\subsection{Middleburry Dataset}\label{sec:middle}
\begin{figure}[htbp]
    \centering
\subfloat[(a)]{\includegraphics[width=0.24\linewidth]{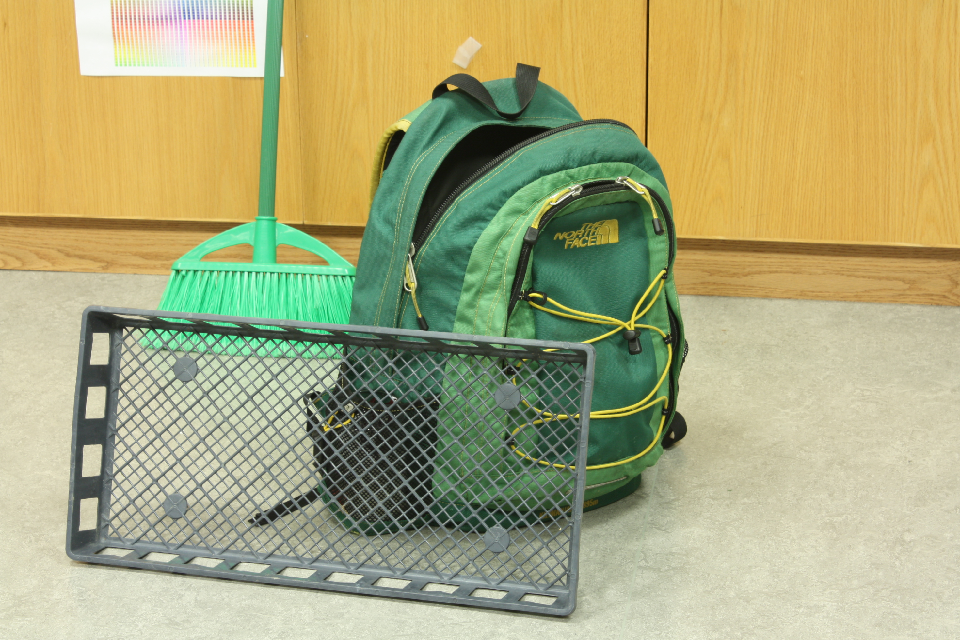}}\hspace{0.1pt} 
\subfloat[(b)]{\includegraphics[width=0.24\linewidth]{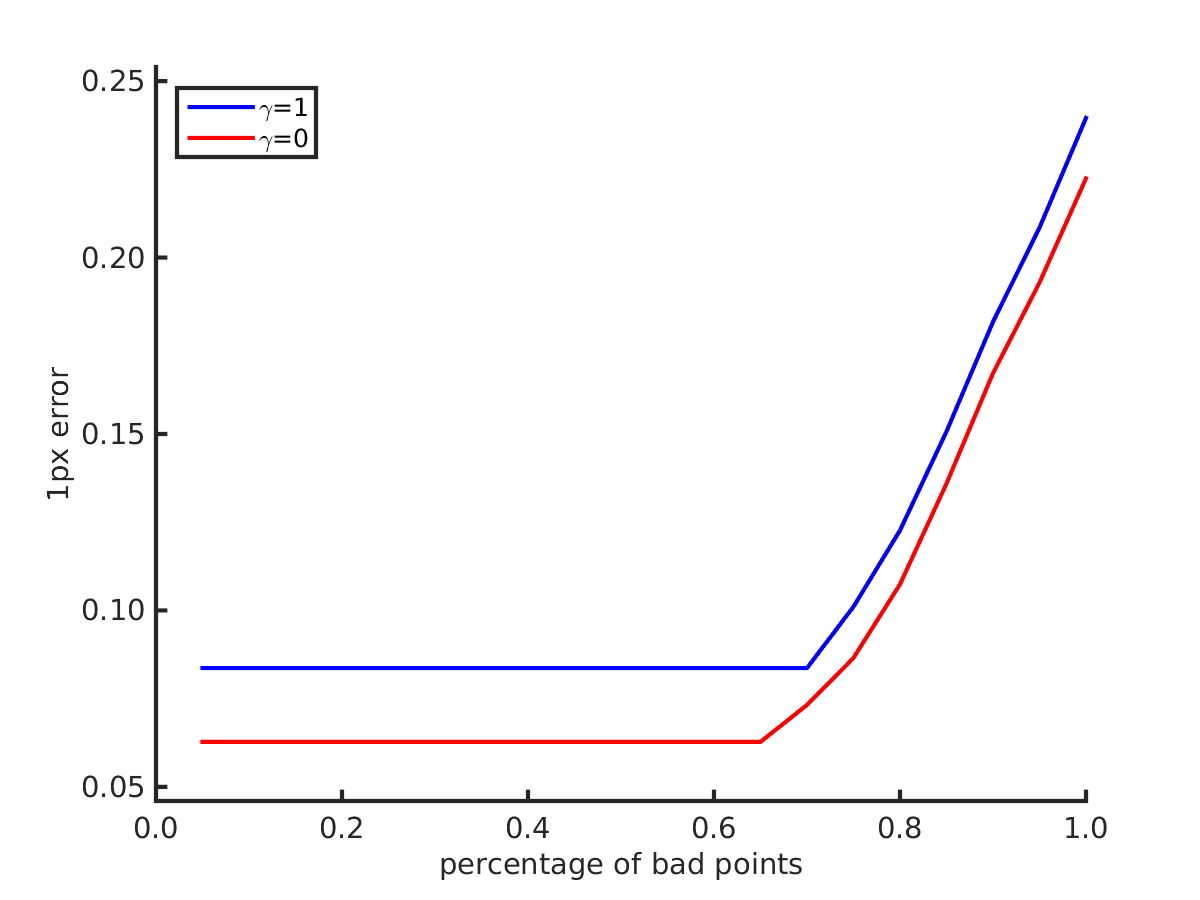}}\hspace{0.1pt}
\subfloat[(c)]{\includegraphics[width=0.24\linewidth]{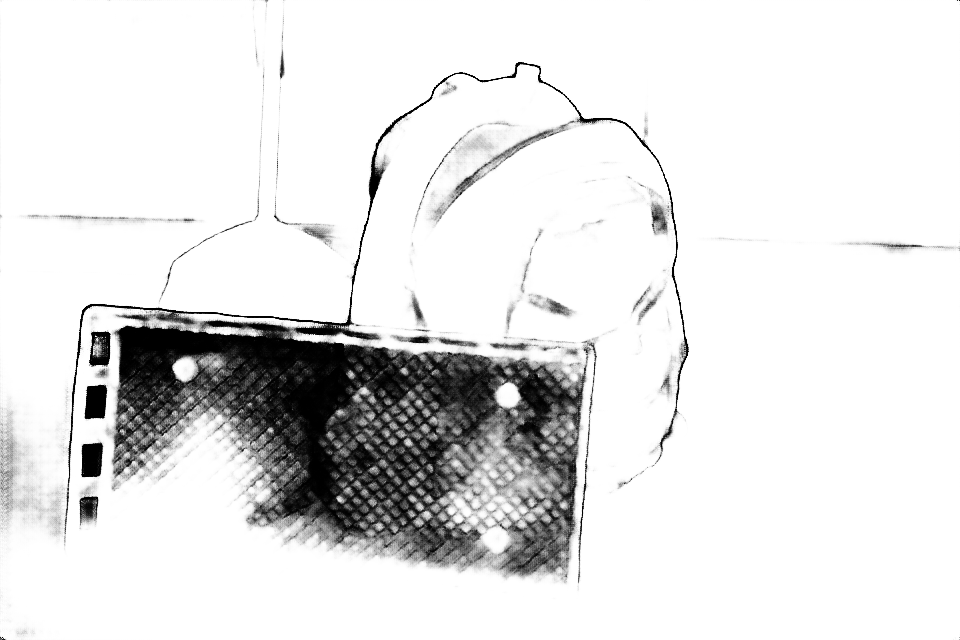}}\hspace{0.1pt}
\subfloat[(d)]{\includegraphics[width=0.24\linewidth]{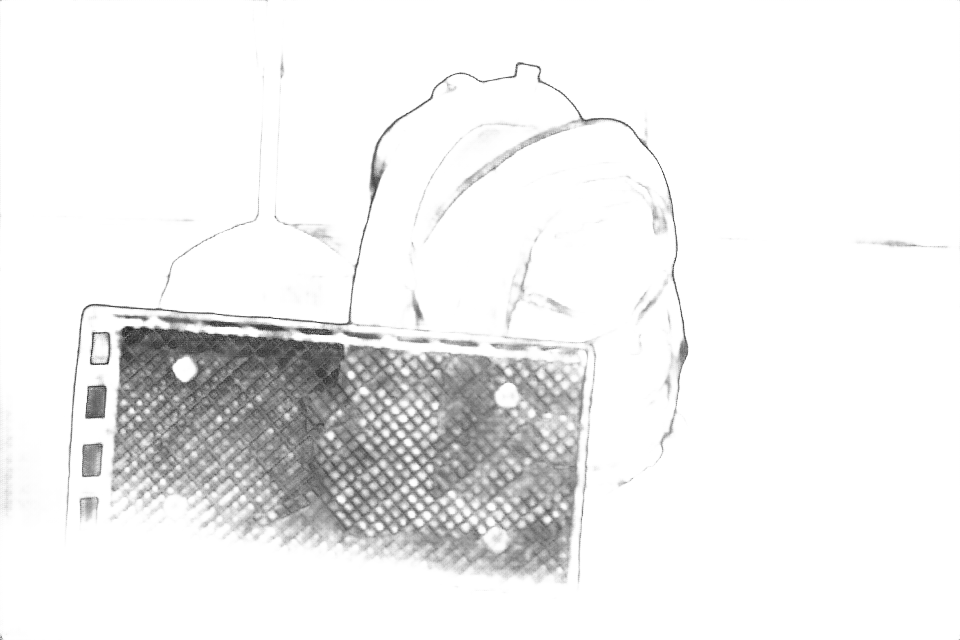}}\hspace{0.1pt}
\subfloat[(e)]{\includegraphics[width=0.24\linewidth]{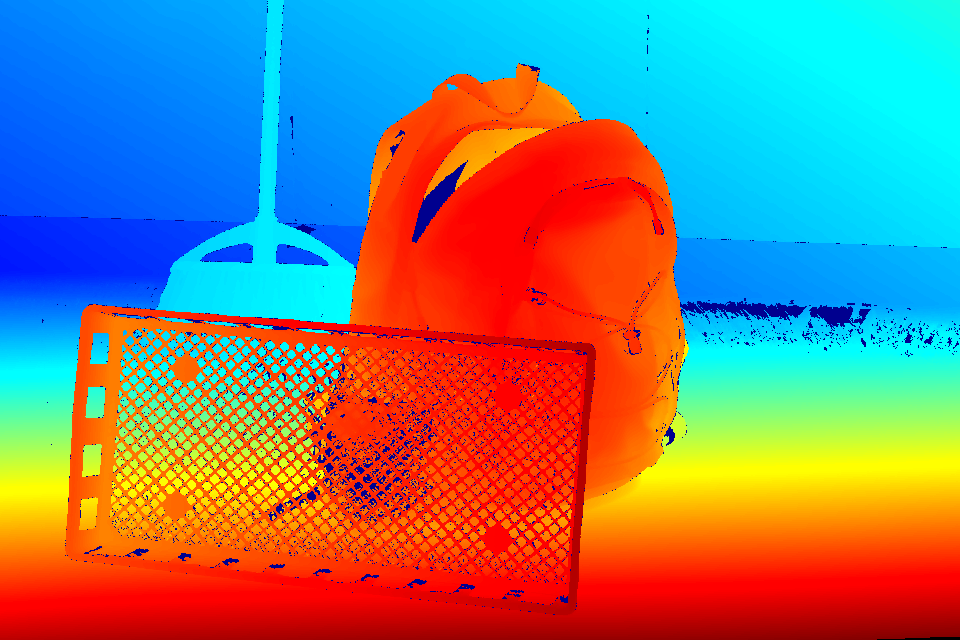}}\hspace{0.1pt} 
\subfloat[(f)]{\includegraphics[width=0.24\linewidth]{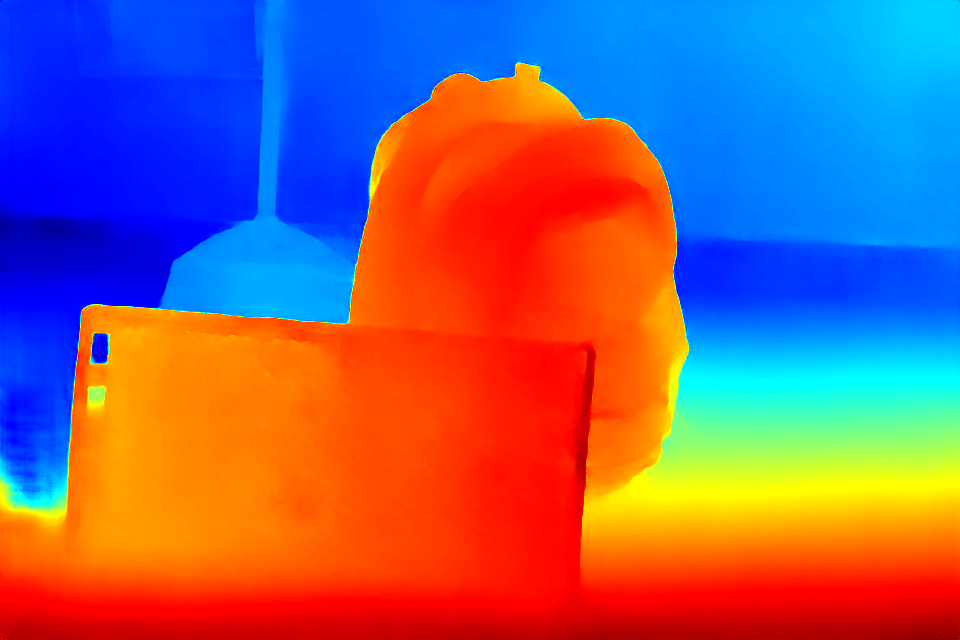}}\hspace{0.1pt}
\subfloat[(g)]{\includegraphics[width=0.24\linewidth]{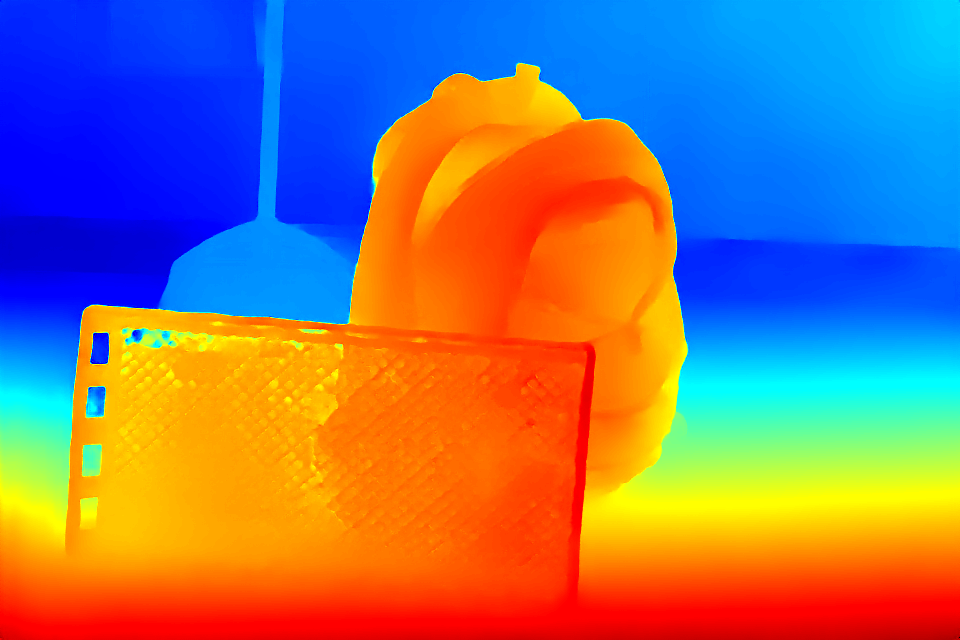}}\hspace{0.1pt}
\subfloat[(h)]{\includegraphics[width=0.24\linewidth]{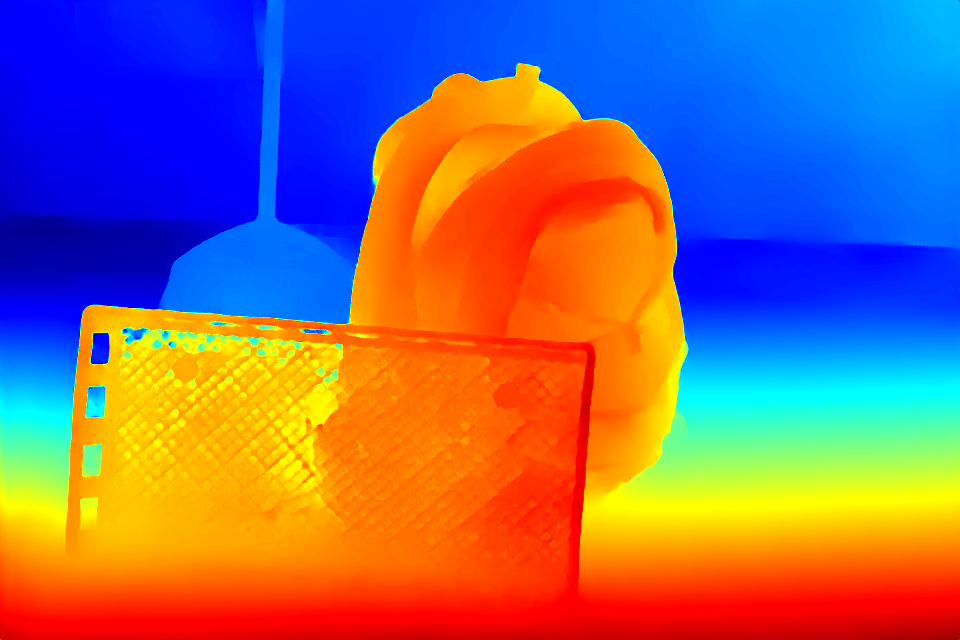}}\hspace{0.1pt} 
\setcounter{subfigure}{0}
\vspace{-3pt}
 \caption{
 Some visual results on Middleburry 2014~\cite{scharstein2014high} dataset. (a) Reference image; (b) ROC curve for $\gamma=0$ (red) and $\gamma=1$ (blue); (c)-(d): Confidence map for $\gamma=0$ and $\gamma=1$, respectively. Dark region means low confidence; (e)-(h): Disparity maps of ground-truth, DispFullNet, our method with $\gamma=0$, and our method with $\gamma=1$, respectively. Notice that the confidence map is able to capture large-error regions, such as the grid. Meanwhile our method produces better disparity prediction, such as the holes on the left of the image. Best view in color. 
}
 \label{fig:mid}
\end{figure}

\begin{figure}[htbp]
    \centering
{\includegraphics[width=0.8\linewidth]{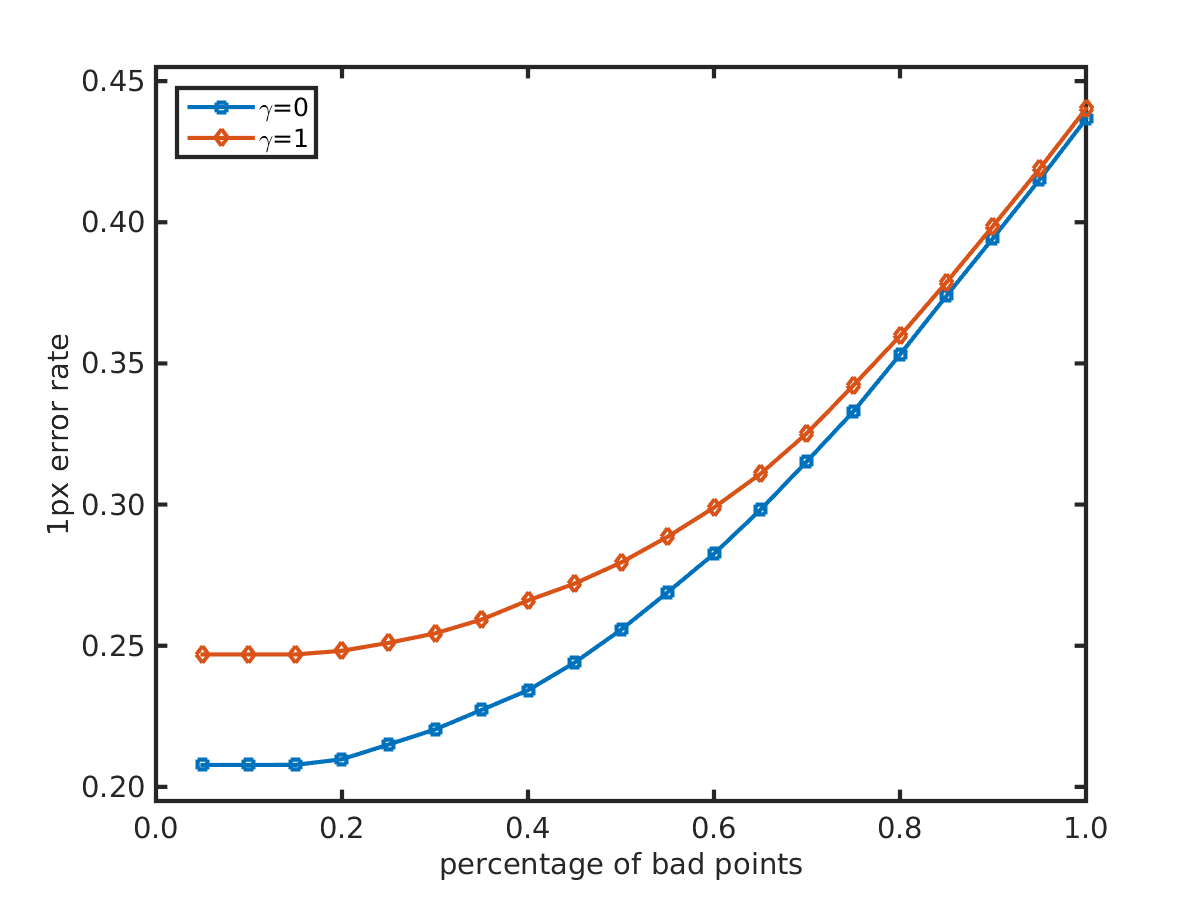}}\hspace{0.5pt} 
 \caption{ROC curve for $\gamma=0$ and $\gamma=1$ on the Middlebury dataset. We observe that ROC of $\gamma=0$ stays below the ROC of $\gamma=1$ all the time, which suggests better disparity prediction and confidence learning.}
 \label{fig:rocmid}
\end{figure}

We use Middleburry 2014~\cite{scharstein2014high} dataset to test the generalization ability of our approach. We use model trained in Section ~\ref{sub:abl} directly without further fine-tuning. Considering search range of the models and computation capacity, we resize the images to $960 \times 640$ in our evaluations. The results are summarized in Table.~\ref{tab:mid}. The ROC curves for $\gamma=0$ and $\gamma=1$ are presented in Fig.~\ref{fig:rocmid}. Compared to original DispFullNet model, it can be observed that our approach achieves significantly lower error in terms of EPE and error rate. It indicates that our confidence learning approach reduces the over-fitting to the given dataset and thus has better generalization ability. Meanwhile, notice that $\gamma=0$ is better than $\gamma=1$ on both of confidence measure (higher ratio) and disparity estimation (lower EPE). It indicates that the $\gamma=0$ model is more robust. Some visual results are presented in Fig.~\ref{fig:mid}. A possible explanation is that by focusing more on normal regions, the CNN is able to learn with less perturbation from noise, such as occlusion and texture-less regions, which results in better generalization when the model is directly deployed to a different domain without any domain adaption techniques, such as fine-tuning.

\subsection{KITTI Dataset}\label{kitti}
We applied our approach on the KITTI Stereo 2015~\cite{Menze2015CVPR} Dataset as well. Note that the ground-truth of the dataset is sparse. There is no ground-truth information in some areas, such as edges of objects and occlusion regions, which is not preferred for confidence evaluating and training, as the important regions for confidence learning are greatly missing. Meanwhile, the fact that ground-truth is sparse prevent us from making evaluation on confidence map, thus only disparity is assessed in this section. 
In detail, We deploy the models pre-trained on solely FlyingThings3D~\cite{mayer2016large} Dataset in two ways, without fine-tuning and with fine-tuning. \\

\begin{table*}[]
\centering
\begin{tabular}{|c|c|c|ccccccc|c|}
\hline
\multirow{3}{*}{Method}                                                & \multirow{3}{*}{\begin{tabular}[c]{@{}c@{}}Conf-guided\\ Ensemble\end{tabular}} & \multirow{3}{*}{$\gamma$} & \multicolumn{2}{c|}{\textbf{Without fine-tuning}}        & \multicolumn{5}{c|}{\textbf{With fine-tuning}}                                                                \\ \cline{4-11} 
                                                                       &                                                                                 &                        & \multicolumn{2}{c|}{KITTI Val} & \multicolumn{2}{c|}{KITTI Val}                      & \multicolumn{3}{c|}{KITTI Test}                \\
                                                                       &                                                                                 &                        & EPE            & \multicolumn{1}{c|}{D1-all}         & EPE            & \multicolumn{1}{c|}{D1-all}        & D1-bg         & D1-fg         & D1-all        \\ \hline
\begin{tabular}[c]{@{}c@{}}DispFulNet~\cite{pang2017cascade}\end{tabular}                 & -                                                                               & -                       & 1.512          & \multicolumn{1}{c|}{10.14}         & 0.727          & \multicolumn{1}{c|}{2.34}          & 3.25          & \textbf{4.21} & 3.41          \\ \hline
\multirow{4}{*}{\begin{tabular}[c]{@{}c@{}}Our \\ Method\end{tabular}} & \multirow{2}{*}{No}                                                             & 0                      & 1.440          & \multicolumn{1}{c|}{8.38}          & 0.703          & \multicolumn{1}{c|}{2.23}          & 3.01          & 6.78          & 3.65          \\
                                                                       &                                                                                 & 1                      & 1.466          & \multicolumn{1}{c|}{8.76}          & 0.695          & \multicolumn{1}{c|}{2.22}          & 2.88          & 5.96          & 3.39          \\ \cline{2-11} 
                                                                       & \multirow{2}{*}{Yes}                                                            & 0                      & \textbf{1.313} & \multicolumn{1}{c|}{\textbf{7.68}} & 0.671          & \multicolumn{1}{c|}{2.01}          & 2.93          & 4.97          & 3.27          \\
                                                                       &                                                                                 & 1                      & 1.331          & \multicolumn{1}{c|}{7.84}          & \textbf{0.668} & \multicolumn{1}{c|}{\textbf{2.00}} & \textbf{2.83} & 4.64          & \textbf{3.13} \\ \hline
\end{tabular}
\vspace{3pt}
\caption{Experimentation result on KITTI 2015~\cite{Menze2015CVPR} Dataset. The best results for each column are bold.}
\label{tab:kitti_tab}
\end{table*}

\subsubsection{Confidence-guided ensemble scheme}
Theoretically, our approach enforce the model to focus more on high-confidence regions, so it is possible that the model makes bad prediction on low-confidence regions, which downgrades the over-all performance. To overcome the shortcoming and show that our confidence is reasonable, one straight-forward solution is to adopt a confidence-guided ensemble scheme, in which we replace the prediction with low-confidence by corresponding estimation from the baseline model (DispFullNet~\cite{pang2017cascade}). Specifically, the disparities of the least $15\%$ confidence are replaced in our experimentation. The evaluations are summarized in Table.~\ref{tab:kitti_tab}

\subsubsection{Experiment observation}
\begin{figure}[htbp]
    \centering
{\includegraphics[width=0.95\linewidth]{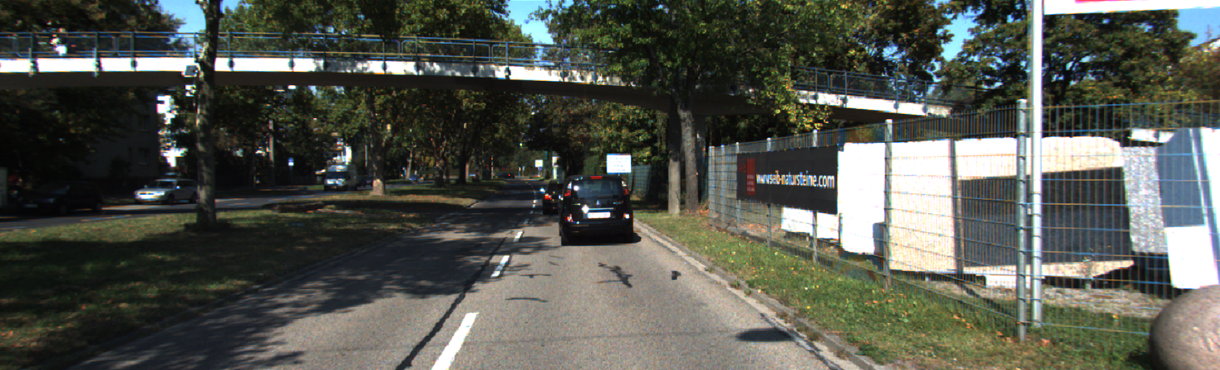}}\hspace{0.5pt} 
{\includegraphics[width=0.95\linewidth]{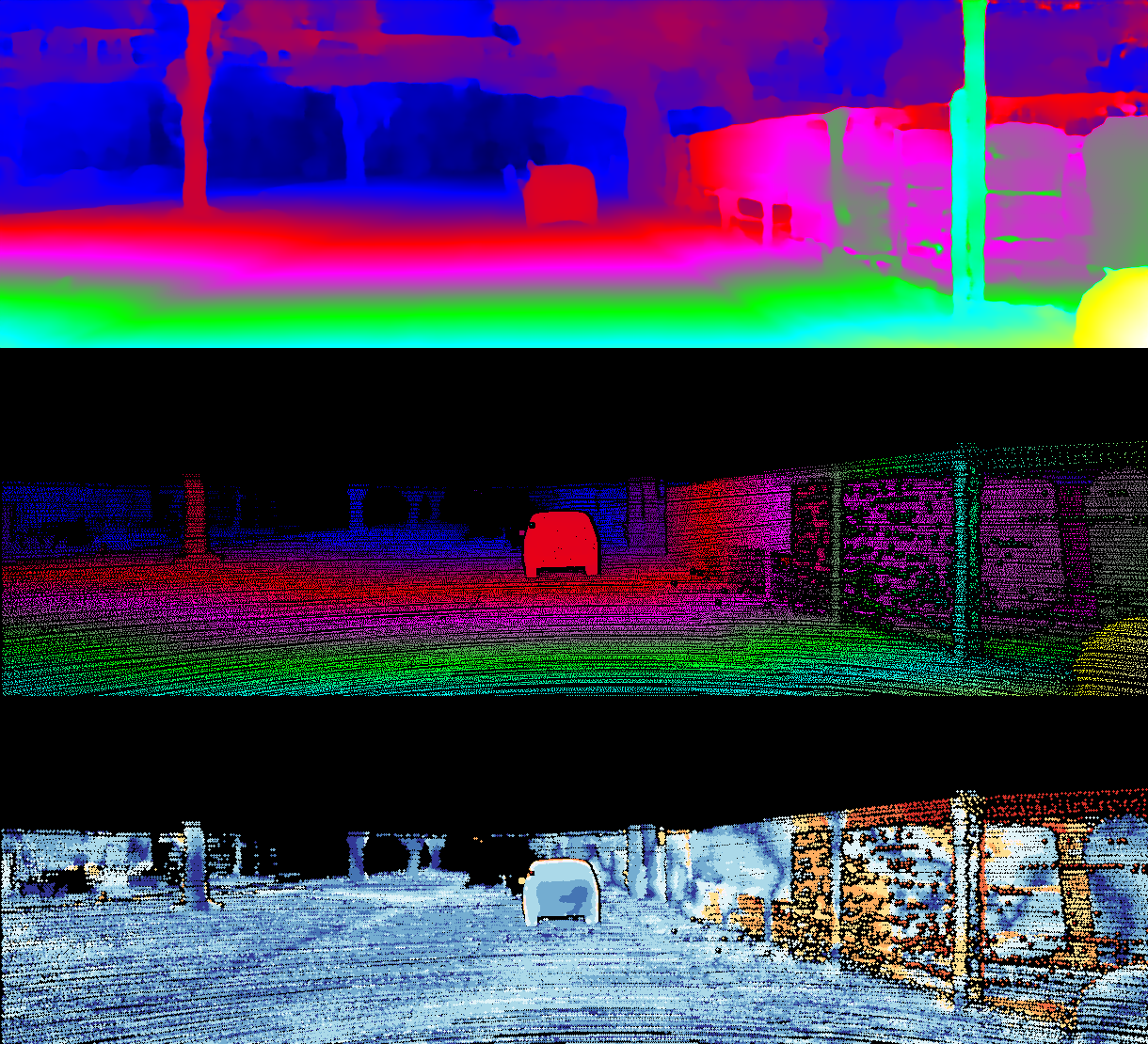}}\hspace{0.5pt} 
{\includegraphics[width=0.95\linewidth]{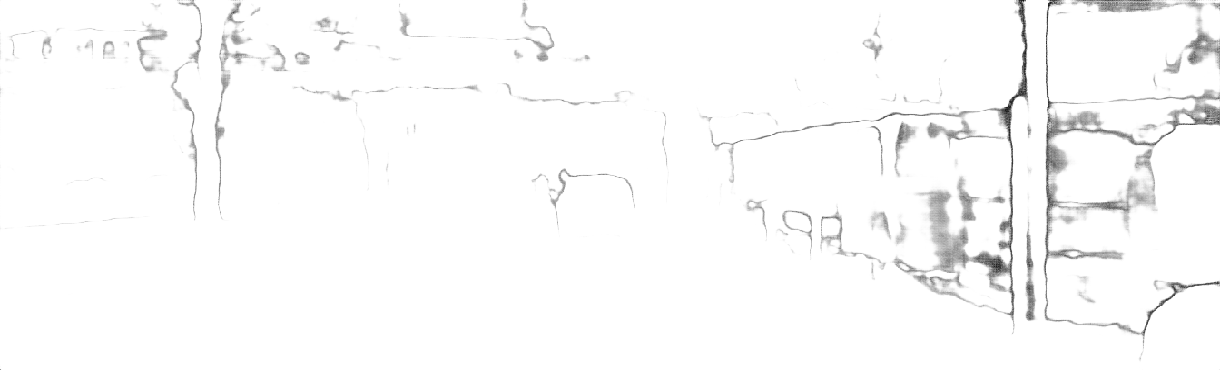}}\hspace{0.5pt} 
 \caption{A visual example of KITTI 2015 validation dataset for $\gamma=1$. The first to the fifth row refers to the reference image, estimated disparity map (not conf-guided ensembled), ground-truth disparity map, 3-pixel error map and estimated confidence map, respectively. Dark regions mean low confidence in the confidence maps. We can observe that low confidence regions are able to match the regions with large disparity prediction error. Best view in color.
}
 \label{fig:kitti}
\end{figure}

\textbf{Without fine-tuning:} The model is directly deployed to the validation dataset. The input image is resized to $1280 \times 375$. As a result, Our method produces better results than DispFullNet~\cite{pang2017cascade} without fine-tuning. Meanwhile, as shown in "Without fine-tuning" column of Table.~\ref{tab:kitti_tab}, $\gamma=0$ outperforms $\gamma=1$ sightly, which is consistent with our observations in Middleburry~\cite{scharstein2014high} Dataset in Sec.~\ref{sec:middle}.\\
\textbf{With fine-tuning:} For fine-tuning experimentation, we set learning rate to $5*10^{-5}$ for the first $50$k iterations, then $1*10^{-5}$ for another $10$k iterations. The input images are randomly cropped to $1216 \times 320$ as data augmentation and a batch size of 6 during training. In the testing stage, the images are resized to $1280 \times 375$. A visual example for $\gamma=1$ is presented in Fig.~\ref{fig:kitti}. Interestingly, after fine-tuning, $\gamma=1$ became better than $\gamma=0$. And in the test dataset, $\gamma=0$ is even outperformed by DispFullNet~\cite{pang2017cascade}. It supports our formulation on confidence in Sec.~\ref{sec:conf_learning}. To train a good model, A proper value of $\gamma$ is significant to determine to what extent the aggressive gradients should be diminished on hard regions.\\
\textbf{With confidence-guided ensemble scheme:} By adopting the confidence-guided ensemble scheme mentioned above, evident decrease is observed in EPE and D1-all in all cases. It suggests that our method locates reasonable low-confidence regions and meanwhile predicts better disparity on the high confidence regions.

\section{Discussion and Conclusion}
\label{sec:conclude}
In this work, we propose a confidence inference method with probabilistic interpretation. We show that proper confidence can be inferred both analytically and experimentally. At the same time, the model can reach an even better convergence state. The inferred confidence can be employed to facilitate the decision-making or the post-processing tasks. Though the proposed method is applied in stereo matching, we believe the same theory can be helpful and extended to other regression problem. 



{
\bibliographystyle{ieee}
\bibliography{ref}
}

\end{document}